\documentclass[review]{elsarticle}

\usepackage{lineno,hyperref}
\usepackage{subfigure}
\usepackage{algorithmicx,algorithm,algpseudocode}
\usepackage{verbatim}
\journal{Journal of \LaTeX\ Templates}
 
\begin{document} 

\begin{frontmatter}
\title{Residual Objectness for Imbalance Reduction} 
\author[ustc_bdaa]{\small{Joya Chen}}\ead{chenjoya@mail.ustc.edu.cn}
\author[ustc_geo]{Dong Liu}\ead{dongeliu@ustc.edu.cn}
\author[tencent]{Bin Luo}\ead{luobinluo@tencent.com}
\author[tencent]{Xuezheng Peng} \ead{reuspeng@tencent.com}
\author[ustc_bdaa]{Tong Xu\corref{correspondingauthor}}\ead{tongxu@ustc.edu.cn}
\author[ustc_bdaa]{Enhong Chen}\ead{cheneh@ustc.edu.cn}
\cortext[correspondingauthor]{\small{Corresponding author}}
\address[ustc_bdaa]{Anhui Province Key Laboratory of Big Data Analysis and Application, University of Science and Technology of China}
\address[ustc_geo]{CAS Key Laboratory of Technology in Geo-Spatial Information Processing and Application System, University of Science and Technology of China}
\address[tencent]{Tencent TaiQ Team}
\begin{abstract} 
For a long time, object detectors have suffered from extreme imbalance between foregrounds and backgrounds. 
While several sampling/reweighting schemes have been explored to alleviate the imbalance, they are usually heuristic and demand laborious hyper-parameters tuning, which is hard to achieve the optimality.  
In this paper, we first reveal that such the imbalance could be addressed in a learning-based manner.
Guided by this illuminating observation, we propose a novel \emph{Residual Objectness (ResObj)} mechanism that addresses the imbalance by end-to-end optimization, while no further hand-crafted sampling/reweighting is required. 
Specifically, by applying multiple cascaded objectness-related modules with residual connections, we formulate an elegant consecutive refinement procedure for distinguishing the foregrounds from backgrounds, thereby progressively addressing the imbalance. 
Extensive experiments present the effectiveness of our method, as well as its compatibility and adaptivity for both region-based and one-stage detectors, namely, 
the \textit{RetinaNet-ResObj}, \textit{YOLOv3-ResObj} and \textit{FasterRCNN-ResObj} achieve relative 3.6\%, 3.9\%, 3.2\% Average Precision (AP) improvements compared with their vanilla models on COCO, respectively.
\end{abstract}
\begin{keyword}
Object Detection \sep Class Imbalance \sep Residual Objectness
\end{keyword} 
\end{frontmatter} 
%\linenumbers

\section{INTRODUCTION}~\label{introduction}

Deep network-based object detectors become prevalent since the success of Region-based CNN (R-CNN)~\cite{rcnn}. 
R-CNN-like detectors~\cite{fast_rcnn,faster_rcnn,rfcn,mask_rcnn,cascade_rcnn,libra_rcnn} work in two stages: 
the proposal stage, e.g., Selective Search~\cite{ss}, EdgeBoxes~\cite{edgebox} and RPN~\cite{faster_rcnn} that samples candidate regions of interest, 
followed by the per-region stage for bounding-box regression and classification.
Meanwhile, one-stage detectors~\cite{yolo,ssd,yolov2,focal_loss,refinedet,yolov3} abandon the per-region stage to pursue higher computational efficiency. 
At most cases, both of them mostly rely on an anchoring mechanism to cover shape-diversity and scale-variant objects, where dense boxes are sampled uniformly over the spatial domain.
Nevertheless, it always causes an extreme foreground-background imbalance as the number of anchors are enormous (e.g. \textasciitilde100\textit{k}), which may result in easy negatives dominated training.
Most recently, anchor-free detectors~\cite{densebox, cornernet, extremenet, garpn, foveabox, centernet} gain much attention due to their simplicity, 
but they are driven by key-point detection (e.g. the center point), thus still suffering from the similar imbalance.

Indeed, not all detectors are equally affected by the foreground-background imbalance. In region-based detectors, such the imbalance has been greatly alleviated by RPN, which rapidly narrows down the huge anchors to a small number, filtering out most negatives. 
For imbalance sensitive one-stage detectors, sampling/reweighting schemes are widely adopted, e.g. Focal Loss~\cite{focal_loss} and GHM~\cite{ghm}. 
Similarly, anchor-free detectors~\cite{cornernet,extremenet,foveabox,centernet} apply Focal Loss or its variants for key-point prediction. 
Despite being effective, these schemes are usually heuristic and demand laborious hyper-parameters tuning. For instance, OHEM~\cite{ohem} only selects hard examples and requires setting mini-batch size with positive-negative proportion, 
whereas Focal Loss~\cite{focal_loss} reshapes the standard cross-entropy loss by two factors to down-weight the well-classified examples.
The GHM~\cite{ghm}, however, hypotheses the very hard examples as outliers and introduce a series of assumptions for gradient harmonizing. 
As illustrated in~\cite{ghm}, it is difficult to define the optimal strategy for addressing the imbalance.

In this work, we explore for object detectors, whether the complicated, heuristic sampling/reweighting schemes could be replaced by a simple, learning-based algorithm.
Our investigation suggests that, with an objectness module \footnote{The objectness module is responsible for distinguishing the foregrounds from the backgrounds, e.g. objectness prior~\cite{yolo,yolov2,yolov3,ron} and anchor refinement module~\cite{refinedet,srn,dsfd}.} substituting for Focal Loss, 
the RetinaNet~\cite{focal_loss} also achieves competitive detection accuracy. In-depth analysis reveals that while the objectness is applied, the extreme imbalance has been implicitly alleviated.
Motivated by this, we present a novel, fully learning-based \emph{Residual Objectness (ResObj)} mechanism, which utilizes multiple cascaded objectness-related modules to address the imbalance, without any hand-crafted sampling/reweighting schemes.
Specifically, we first transfer the imbalance of multi-class classification to the objectness module, to down-weight the contributions of overwhelming backgrounds.
Subsequently, by building residual connections between objectness-related modules, we reformulate the objectness estimation to an elegant consecutive refinement procedure, thereby progressively addressing the imbalance. 

To demonstrate the effectiveness of the proposed mechanism, we incorporate it into the well-known RetinaNet~\cite{focal_loss}, YOLOv3~\cite{yolov3} and Faster R-CNN~\cite{faster_rcnn} detectors. 
On COCO dataset, with residual objectness rather than Focal Loss, the RetinaNet improves 1.3 average precision (AP) than its vanilla model. Meanwhile, by using residual objectness mechanism to 
enhance the original objectness module, the upgraded YOLOv3 surpasses the baseline 1.0 AP. While replacing random sampling with residual objectness for RPN, the Faster R-CNN has achieved 1.2 AP improvement. 
Furthermore, inference speed measurement indicates that the proposed mechanism is very cost efficient.
In conclusion, the residual objectness helps RetinaNet, YOLOv3 and Faster R-CNN to achieve relative 3.6\%, 3.9\%, 3.2\% higher detection accuracy, respectively. But most impressively, benefiting from the fully learning-based architecture, 
our residual objectness rarely requires hyper-parameters tuning to address the foreground-background imbalance in object detection, which has not been explored before.

To summarize, we highlight below the contributions:

$\bullet$ For the first time, we discover that the foreground-background imbalance in object detection could be addressed in a learning-based manner, 
without any hard-crafted sampling/reweighting schemes.

$\bullet$ We propose a novel, fully learning-based \emph{Residual Objectness (ResObj)} mechanism to address the imbalance. 
With a cascade architecture to gradually refine the objectness estimation, our residual objectness is easily optimized end-to-end, and avoiding laborious hyper-parameters tuning as well. 

$\bullet$ We validate the effectiveness of residual objectness on the challenging COCO dataset with thorough ablation studies. 
For various detectors, it steadily improves relative 3\textasciitilde4\% detection accuracy.  

\section{RELATED WORK} 
Our work draws on the recent developments in object detection, class imbalance, objectness estimation and cascaded architecture, which have been discussed as follows.

\noindent\textbf{Classic Object Detectors}.
Sliding window paradigm with hand-crafted features was widely used for object detection. Representatives include Viola and Jones face detector~\cite{jones} and deformable part model (DPM)~\cite{dpm}. 
However, recent years have witnessed the success of CNN-based object detectors, which outperform the classic detectors by a large margin on the benchmarks~\cite{pascal_voc,coco}.

\noindent\textbf{Region-based Detectors}.
Region-based (also called two-stage) object detection is introduced and popularized by R-CNN~\cite{rcnn}. It firstly generates a sparse set of candidate object proposals by some low level vision algorithms~\cite{ss,edgebox}, then determines the accurate bounding boxes and the classes by convolutional networks. A number of R-CNN variations~\cite{fast_rcnn,faster_rcnn,rfcn,mask_rcnn} appear over years, yielding a large improvement in detection accuracy. Among them, Faster R-CNN~\cite{faster_rcnn} is one of the most successful schemes. It introduces region proposal network (RPN), which has became a common module in two-stage approaches.

\noindent\textbf{One-stage Detectors}.
For real time detection, one-stage detectors abandon the usage of the second-stage. Most of them~\cite{ssd,yolov2,focal_loss,refinedet,yolov3} apply a dense set of anchors to to cover shape-diversity and scale-variant objects. RefineDet~\cite{refinedet} can be seen as a combination of two-stage and one-stage, which uses two-step regression for bounding-boxes but avoids per-region stage. 

\noindent\textbf{Objectness}.
Objectness usually represents how likely a box covers an object~\cite{objectness}. Classic objectness methods~\cite{objectnesswhat,objectnessBING} are used to reduce the number of proposal windows for faster detection.  RPN~\cite{faster_rcnn} can also be viewed as an objectness method. There are also one-stage detectors using objectness-like mechanism to alleviate the foreground-background class imbalance, including objectness prior in RON~\cite{ron}, objectness score in YOLO~\cite{yolo,yolov2,yolov3} and anchor refinement module in RefineDet~\cite{refinedet}. Recent state-of-the-art one-stage face detectors~\cite{srn,dsfd} also adopt the module.

\noindent\textbf{Class Imbalance}.
Recent deep anchor-based detectors often face an extreme foreground-background class imbalance during training. 
As the region-based detectors have proposal stage, the one-stage detectors are more imbalance sensitive than region-based ones. 
Previous methods for handling class imbalance can be divided into two categories: 
(1) Hand-crafted sampling/reweighting schemes, including random sampling~\cite{faster_rcnn}, OHEM~\cite{ssd,ohem}, Focal Loss~\cite{focal_loss}, GHM~\cite{ghm} and IoU-balanced sampling~\cite{libra_rcnn}. 
(2) Introduction of objectness module~\cite{yolo,yolov2,refinedet,yolov3,ron,srn,dsfd,probj}, which can be seen as an imitation of the region-based detectors. 
But most detectors with objectness module~\cite{yolo,yolov2,refinedet,ron,srn,dsfd} still maintain sampling heuristics or hard example mining schemes.  

\noindent\textbf{Cascaded Architectures}.
There have been several attempts~\cite{cascade_rcnn,craft,gcnn,deepproposals,ilr,crpn} that apply cascade architecture to reject easy samples at early layers or stages, and regress bounding boxes iteratively for progressive refinement.
However, none of them are designed for one-stage detectors.

\noindent\textbf{Comparison and Difference}.
We summarize the differences between the residual objectness and previous works from there aspects. 

$\bullet$ Class imbalance: For object detection, the sampling/reweighting schemes are widely adopted~\cite{faster_rcnn,libra_rcnn,ssd,focal_loss,ohem,ghm} to alleviate the imbalance between foregrounds and backgrounds.  
Nevertheless, they are usually heuristic and costly to tune. Conversely, our residual objectness addresses the imbalance in fully learning-based manner, which is easily optimized end-to-end.

$\bullet$ Objectness: This RPN-like module has been applied in many detectors~\cite{yolo,yolov2,refinedet,yolov3,ron,srn,dsfd,probj}. 
Initially, it is supposed to distinguish the foregrounds from the backgrounds. Nevertheless, such the module usually suffers from extreme foreground-background imbalance, 
thus only be used for coarse selection in most cases. Instead of reusing the hard-crafted sampling/reweighting schemes like~\cite{yolo,yolov2,refinedet,ron,srn,dsfd}, 
we reformulate the objectness estimation to a cascaded refinement procedure, thereby progressively addressing the imbalance in a learning-based manner. 

$\bullet$ Cascaded Architectures: our residual objectness is supposed to progressively address the foreground-background imbalance, 
which is similar with recent cascaded architectures~\cite{cascade_rcnn,craft,gcnn,deepproposals,ilr,crpn} progressively refine bounding-boxes. 
However, most of them~\cite{cascade_rcnn,craft,gcnn,deepproposals,ilr} are only applicable for the per-region stage, whereas the only exception C-RPN~\cite{crpn} is designed for object tracking. 
Our proposed mechanism is generalized for both region-based and one-stage detectors. 

\section{METHODOLOGY} 
In this section, we first discuss our motivation, why learning-based approaches should be applied to address the foreground-background imbalances. 
Based on it, we describe our learning-based algorithm Residual Objectness, which addresses the imbalance by end-to-end optimization without any hand-crafted sampling/reweighting schemes. 
\subsection{Motivation}\label{motivation}
We introduce our motivation starting from revisiting the limitations of existing solutions for addressing the foreground-background imbalance.
The first camp --- hand-crafted sampling/reweighting schemes~\cite{faster_rcnn,libra_rcnn,ssd,focal_loss,ohem,ghm}, always introduces network-independent hyper-parameters that require laboriously tuning, which is hard to achieve the optimality.
Instead, the second camp --- learning-based algorithms~\cite{yolo,yolov2,refinedet,yolov3,ron,srn,dsfd,probj} could remove easy negatives by end-to-end optimization, but still have an extreme imbalance in the RPN-like/objectness module. 
As a result, most of them~\cite{yolo,yolov2,refinedet,ron,srn,dsfd} still maintain the sampling/reweighting schemes. 

Therefore, we manage to replace the complicated, heuristic sampling or reweighting schemes with a simple, fully learning-based algorithms. 
In the following investigation, we replace the well-known Focal Loss with an objectness module, rather than any sampling heuristics or hard example mining schemes.
Specifically, we follow the configuration~\footnote{\raggedright\url{https://github.com/facebookresearch/Detectron/blob/master/configs/12_2017_baselines/retinanet_R-50-FPN_1x.yaml}} of RetinaNet with ResNet-50-FPN~\cite{resnet,fpn} in \texttt{Detectron}~\cite{detectron}, 
using a 600 pixel train and test image scale to keep consistency with the original literature~\cite{focal_loss}.

\begin{figure}[h]  
    \centering 
    \includegraphics[width=\linewidth]{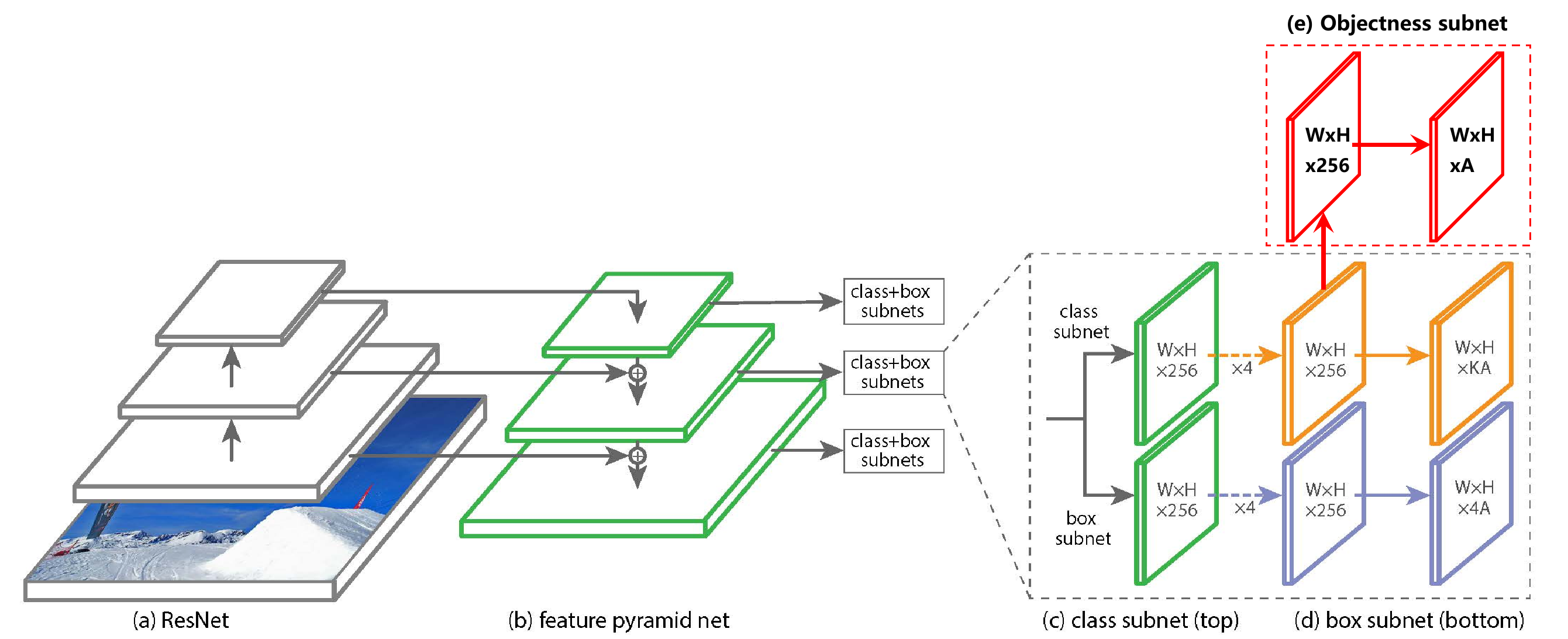}
    \caption{Architecture of RetinaNet with objectness subnet (\textit{RetinaNet-Obj}). 
    The class subnet (c) predicts the class-specific confidence score at each spatial position for \textit{A} anchors and \textit{K} categories, 
    whereas (e) estimates objectness score for each anchors. }\label{figure1} 
\end{figure}

\noindent\textbf{RetinaNet with Objectness.}
As shown in Figure~\ref{figure1}, we simply build two convolutional layers on the top of class subnet to estimate objectness score for each anchor. 
Here, we name this new detector as \textit{RetinaNet-Obj} (\textit{RetinaNet-Objectness}), while the original as \textit{RetinaNet-FL} (\textit{RetinaNet-Focal Loss}). 
To abandon the usage of Focal Loss, during training, the class subnet of \textit{RetinaNet-Obj} alternatively uses cross entropy (CE) loss with sigmoid activation. 
Following previous literatures~\cite{yolo,yolov2,refinedet,yolov3,ron,srn,dsfd,probj}, the objectness module is also trained with binary CE loss with sigmoid activation. 
For keep consistency with vanilla RetinaNet, the criteria of anchor assignment is not changed, where anchors are assigned to ground-truth object boxes using an intersection-over-union (IoU) threshold of 0.5; 
and to background if their IoU is in [0, 0.4). To avoid sampling/reweighting schemes, if an anchor is not assigned to a ground-truth object, it incurs no loss for box and class predictions, but only for objectness.
During inference, we compute the class-specific confidence score by $P(Class) = P(Class|Obj) \times P(Obj)$.
With these preliminaries, we train \textit{RetinaNet-Obj} on COCO \texttt{train2017} and validate it on \texttt{minival}. 

However, at the beginning of training, the large number of background examples will generate a large, destabilizing loss value. To prevent it happening,
we use the biased initialization in ~\cite{focal_loss} to ensure the initial class-specific score and objectness score are \textasciitilde$0.01$ and \textasciitilde$1 / K$, respectively. 

\noindent\textbf{Comparing \textit{RetinaNet-ResObj} with \textit{RetinaNet-FL}.}
After introducing objectness module, the extreme foreground-background imbalance of class subnet is transferred to the objectness subnet. 
In practice, this often amounts to enumerating \textasciitilde100\textit{k} negative cases but merely \textasciitilde100 positives for an image. 
Nevertheless, such the imbalance does not seem to prevent the model from achieving competitive detection accuracy. 
As shown in Table~\ref{table1}, with appropriate thresholds, the \textit{RetinaNet-FL} and \textit{RetinaNet-Obj} have achieved similar results (34.1 AP vs. 34.2 AP).
On the other hand, the vanilla RetinaNet without Focal Loss yields obviously lower 30.2 AP~\footnote{This result is reported while $\gamma=0, \alpha=0.5$ for Focal Loss.}.

\begin{figure}[h]
    \centering
    \includegraphics[width=\linewidth]{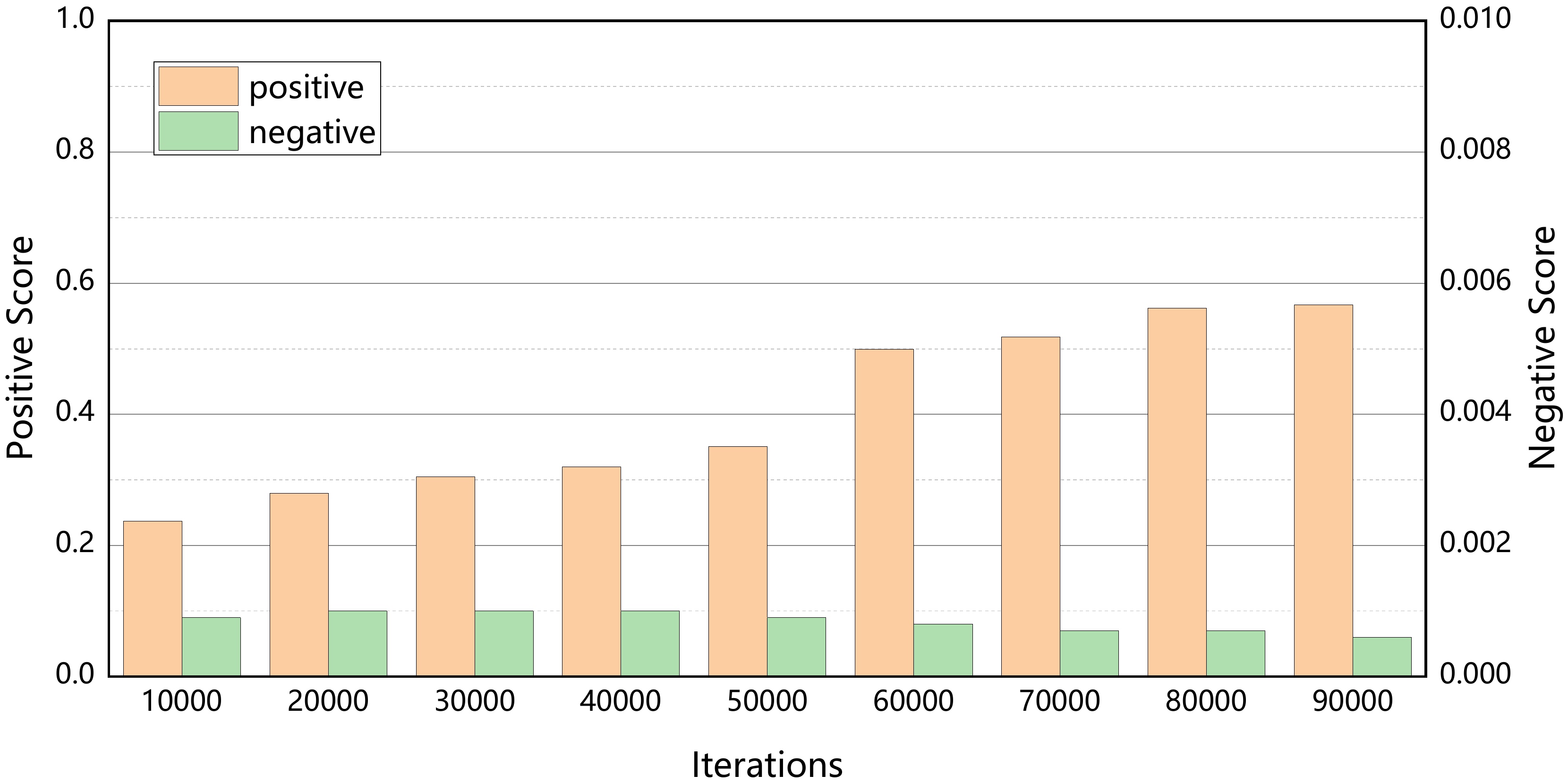} 
    \caption{Average objectness scores on COCO \texttt{minival} vary with training iterations.}\label{figure2}   
\end{figure}
 
\begin{table*}
    \centering
    \scriptsize
    \subtable[Varying threshold for \textit{RetinaNet-FL}]{
        \begin{tabular}[t]{c|c|ccc}
            \hline
            Inference & NMS & AP & AP$_{50}$ & AP$_{75}$ \\ 
            \hline
            0.100 & 0.45 & 34.0 & 53.0 & 36.1 \\ 
            0.050 & 0.50 & \textbf{34.2} & \textbf{53.1} & \textbf{36.5} \\ 
            0.010 & 0.50 & \textbf{34.2} & \textbf{53.1} & \textbf{36.5} \\ 
            0.005 & 0.50 & \textbf{34.2} & \textbf{53.1} & \textbf{36.5} \\ 
            0.001 & 0.50 & \textbf{34.2} & \textbf{53.1} & \textbf{36.5} \\ 
            \hline
        \end{tabular}
    }\label{table1.1}
    \subtable[Varying threshold for \textit{RetinaNet-Obj}]{
        \begin{tabular}[t]{c|c|ccc} 
            \hline
            Inference & NMS & AP & AP$_{50}$ & AP$_{75}$ \\ 
            \hline
            0.100 & 0.45 & 33.1 & 51.8 & 35.4 \\ 
            0.050 & 0.45 & 33.6 & 52.2 & 36.0 \\ 
            0.010 & 0.45 & 33.9 & 52.6 & 36.3 \\ 
            0.005 & 0.50 & 34.0 & 52.5 & \textbf{36.7} \\ 
            \textbf{0.001} & \textbf{0.45} & \textbf{34.1} & \textbf{52.9} & 36.6 \\ 
            \hline
        \end{tabular}
    }\label{table1.2}
    \caption{Varying inference threshold (w. optimal NMS) to evaluate detection accuracy on COCO \texttt{minival}. Very similar results can be achieved by \textit{RetinaNet-FL} and \textit{RetinaNet-Obj}.}\label{table1}
\end{table*}

Why the extreme imbalance on objectness module, does not result in a significant performance degradation? 
As presented in the Figure~\ref{figure2}, we plot the positive/negative average objectness score vary with training iterations. 
Due to the imbalance, the positive objectness score is relative low. Ideally, we want the positive score to be a high value (e.g. 0.9), but it is only up to 0.57 here.
Sampling/reweighting schemes could be reused here to improve its estimation, but a fully learning-based algorithm shows more elegant and convenient.

\subsection{Residual Objectness} 
Based on the analysis above, even if the objectness module suffers from an extreme imbalance, the \textit{RetinaNet-Obj} could achieve similar results with \textit{RetinaNet-FL}.
Therefore, it's natural to construct another objectness-related module to further alleviate such the imbalance. 
Now, we introduce residual objectness, a simple, fully learning-based mechanism for addressing the foreground-background imbalance in object detection.

\noindent\textbf{Imbalance Transfer.}
To make it easier to deal with imbalance, the first step of applying our residual objectness mechanism, is to transfer the imbalance of multi-classification in class subnet to the binary-classification objectness.
For instance, the vanilla RetinaNet directly classifies $K + 1$ object categories, where $+ 1$  denotes the background case. 
By residual objectness mechanism, the detector predicts $K$ object categories with a foreground/background classification.
We will demonstrate why this is better.

For anchor $i$, we define $o_i$ for its objectness score and $p_k^i$ for the score of object category $k$ in $K$ categories, with the ground-truth category label $l_i$.
Among all anchors, there are $P$ foreground examples and $N$ background examples. The classification loss $L_{Retina-FL}$ of \textit{RetinaNet-FL} can be written as: 

\begin{equation}
    L_{Retina-FL} = FocalLoss(\sum_{i=1}^{A}(1\{y_i=0\}L_{i}^{neg} + 1\{y_i>0\}L_{i}^{pos}) \label{retina_fl_loss}
\end{equation}

Where the $L_{i}^{neg}$ and $L_{i}^{pos}$ are the corresponding cross-entropy loss functions:

\begin{equation}
    L_{i}^{neg} = -\sum_{k=1}^{K}\log(1 - p_k^i)
\end{equation}

\begin{equation}
    L_{i}^{pos} = -\sum_{k=1}^{K}1\{y_i=k\}\log(p^i_k) + 1\{y_i \neq k\}\log(1 - p^i_k) 
\end{equation}

Due to numerous easy background examples, the accumulative $L_{i}^{neg}$ may dominate the training procedure. Therefore, Focal Loss is applied for down-weighting them. 
As we will show, the objectness module has a similar effect:

\begin{equation}
    L_{Retina-Obj} = -\sum_{i=1}^{A}L_{i}^{obj}+ 1\{y_i>0\}L_{i}^{pos} \label{retina_obj_loss}
\end{equation}

Where $L_{i}^{obj}$ denotes:

\begin{equation}
    L_{i}^{obj} = - 1\{y_i=0\}\log(1 - o_i)- 1\{y_i>0\}\log(o_i)
\end{equation}

In the Equation~\ref{retina_obj_loss}, the term $1\{y_i>0\}L_{i}^{pos}$ is equal to that in Equation~\ref{retina_fl_loss}. 
Consequently, the overall loss of negative examples are: 

\begin{equation}
    L_{Retina-FL}^{neg} = FocalLoss(-\sum_{i=1}^{A}1\{y_i=0\}\sum_{k=1}^{K}\log(1 - p_k^i))
\end{equation}

\begin{equation}
    L_{Retina-Obj}^{neg} = - \sum_{i=1}^{A}1\{y_i=0\}\log(1 - o^i)
\end{equation}

Without Focal Loss~\cite{focal_loss}, the $L_{Retina-FL}^{neg}$ is approximately $K$ times higher than $L_{Retina-Obj}^{neg}$, 
which illustrates that transferring the imbalance to objectness could relieve it to some extent.

\begin{figure}[h] 
    \centering
    \subfigure[\textit{RetinaNet-FL}]{
    \includegraphics[width=0.8\linewidth]{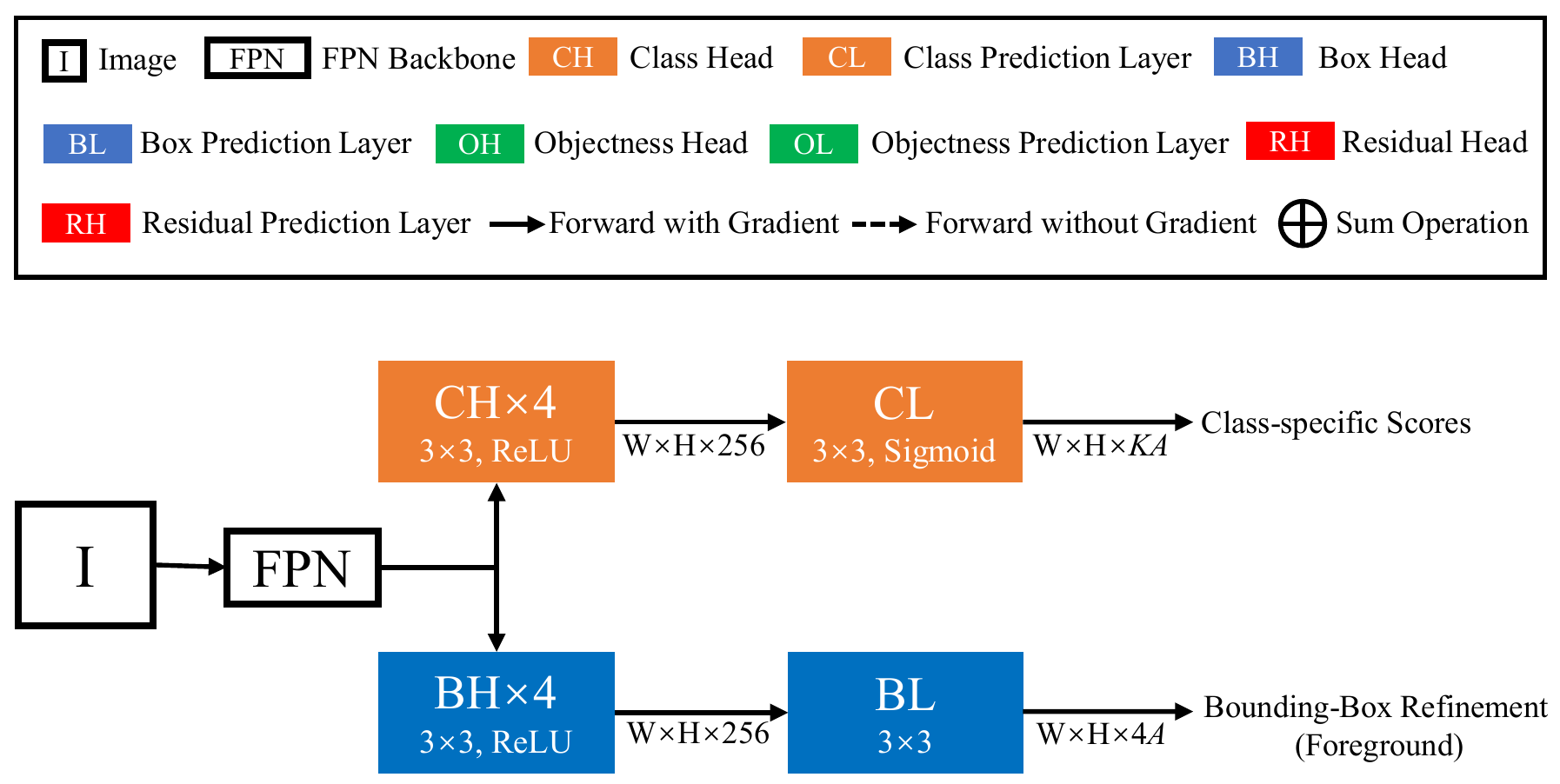}  
    \label{figure3a} 
    }
    \subfigure[\textit{RetinaNet-ResObj}]{
    \includegraphics[width=0.8\linewidth]{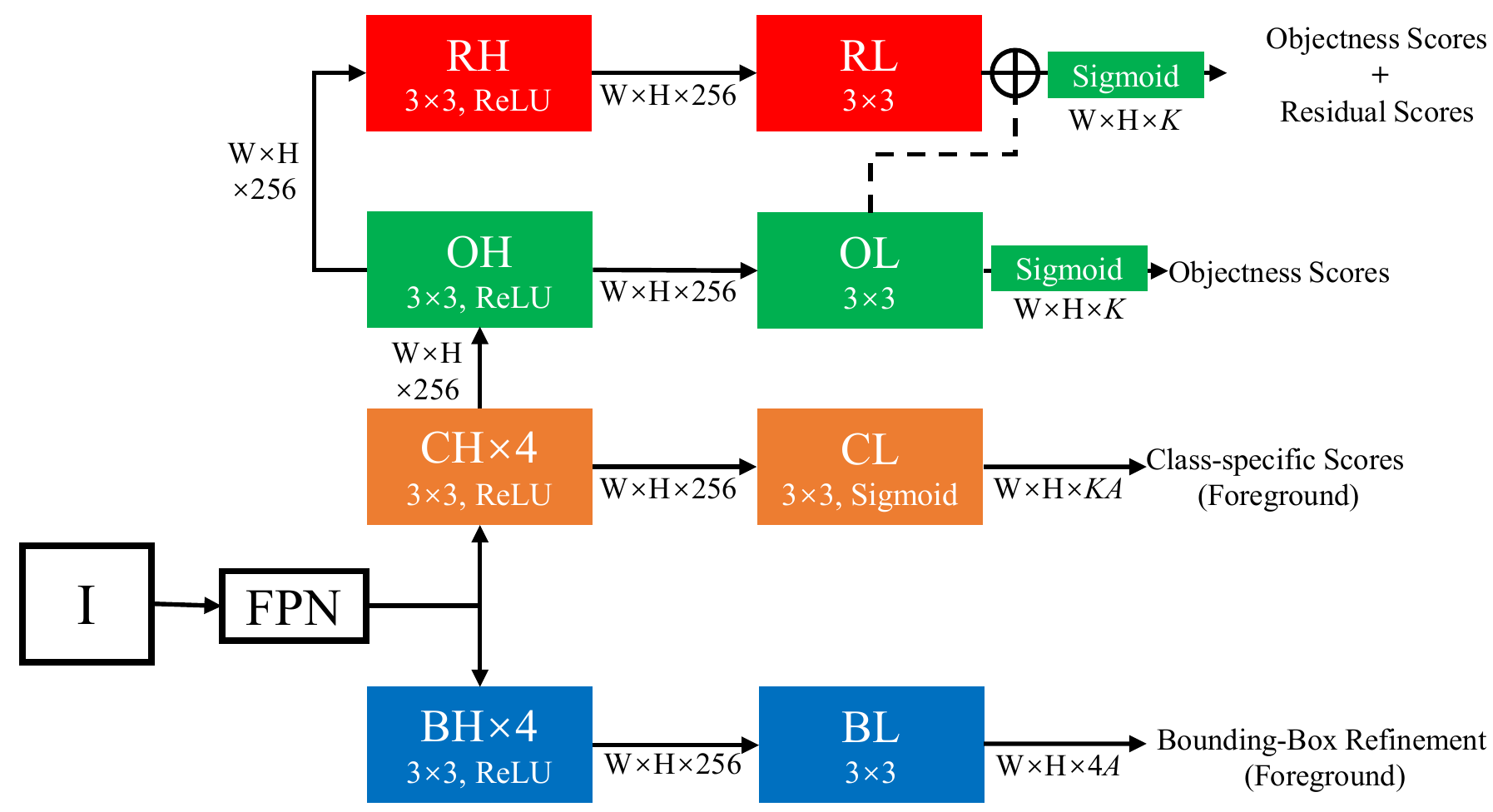}  
    \label{figure3b}
    }
    \caption{Pipeline of \textit{RetinaNet-FL} and \textit{RetinaNet-ResObj}. }\label{figure3}
 \end{figure}

\noindent\textbf{Residual Objectness Architecture.}  
To resolve the foreground-background imbalance of objectness module, we formulate the objectness estimation as $o = o_0 + \sum_{t=1}^{T}r_t$, where $o_0$ denotes initial objectness score, 
and $r_t$ denotes the corresponding refinement in $t$ step. Ideally, the objectness scores for positives could be gradually improved while for negatives reduced. 
To the extreme, if the $o_0$ has been the optimal, each $r_t$ is supposed to be \textasciitilde0. 
We hypothesize that the imbalance is progressively addressed at this procedure.

Firstly, we describe architecture of the vanilla RetinaNet~\cite{focal_loss}. As shown in Figure~\ref{figure3a}, taking an input image, the feature pyramid network (FPN)~\cite{fpn} backbone is responsible for computing a convolutional feature map, 
then for \textit{A} anchors, the class subnet performs classification of \textit{K} object categories, while the box subnet performs bounding-box regression for the foreground anchors.

Now, we introduce our residual objectness architecture. As presented in Figure~\ref{figure3b}, we first build an objectness subnet on the top of class subnet, for transferring the imbalance of classification to it.
Subsequently, multiple residual subnets are followed to refine the objectness estimation. 
To progressively address the imbalance, we perform sum operation between objectness logits and residual logits, with sigmoid activation to compute the updated objectness score.
For encouraging the residual subnet to work independently, we isolate the residual subnet from objectness subnet in backpropagation (dotted line in Figure~\ref{figure3b}).

\begin{algorithm}[h]
    \caption{Residual objectness mechanism during training.}\label{algorithm:training}
    {\bf Input:} \\
    \hspace*{0.23in}Feature produced by class head $f^c$;\\
    \hspace*{0.23in}Objectness subnet $N^o$;\\
    \hspace*{0.23in}Residual step $T$ and subnets $N^r_1, N^r_2, ..., N^r_T$;\\
    \hspace*{0.23in}Ground-truth label $l$; Binary cross entropy loss $BCELoss$;\\
    {\bf Output:} \\
    \hspace*{0.23in}Objectness loss $\mathcal{L}^{obj}$ and Residual loss $\mathcal{L}^{res}$
    \\
    \begin{algorithmic}[1]
    \State Network forward to compute objectness logits $o \gets N^o(f^c)$, residual logits $r_t \gets N_t^r(f^c), t = 1, 2, ..., T$
    \State Objectness Loss $\mathcal{L}^{obj} \gets BCELoss(\sigma(o), l)$, $\sigma()$ is the sigmoid function
    \State $o_{0} \gets o$
    \For{$t=1,...,T$}
        \State Compute minimum positive score $o_{t-1}^{minp}$
        \State $o_t \gets o_{t-1} + r_t$ for the $o_{t-1} \geq o_{t-1}^{minp}$ cases
        \State Residual Loss $\mathcal{L}^{res}_t \gets BCELoss(\sigma(o_t), l)$ for the $o_{t-1} \geq o_{t-1}^{minp}$ cases
    \EndFor
    \State $\mathcal{L}^{res} = \sum_{t=1}^{T}\mathcal{L}^{res}_t$
    \State \Return $\mathcal{L}^{obj}, \mathcal{L}^{res}$
    \end{algorithmic}
\end{algorithm}

\noindent\textbf{Training and Inference.}
Given the residual objectness architecture, we further describe its training and inference. As presented in Algorithm~\ref{algorithm:training}, for each step $t$, 
we refine the objectness prediction at $t-1$ step for the cases above the minimum positive score, which contains all positive cases and not well classified negatives.
While for inference, we simply use the objectness prediction in the final step, and computes class-specific score by $P(Class) = P(Class|Obj) \times P(Obj)$.
In this way, we have successfully modeled a fully learning-based mechanism that could progressively address the imbalance.

\section{EXPERIMENTS}
In this section, we present experimental results of our residual objectness on the challenging COCO~\cite{coco} datasets. 
We first give the description for the dataset, then present implementation details of RetinaNet~\cite{focal_loss}, YOLOv3~\cite{yolov3} and Faster R-CNN~\cite{faster_rcnn} with the residual objectness. 
In Section~\ref{section:ablations}, we conduct thorough ablation experiments to validate the proposed mechanism. 
Finally, Section~\ref{section:results} compares the results of \textit{RetinaNet-ResObj}, \textit{YOLOv3-ResObj} and \textit{FasterRCNN-ResObj} with their vanilla models on the COCO \texttt{test-dev}.

\subsection{Implementation Details}\label{subsection:4.1}
\noindent\textbf{COCO.}
By standard practices~\cite{mask_rcnn,cascade_rcnn,focal_loss}, we adopt COCO \texttt{train2017} for training, which is an union of 80k training images in COCO \textit{train2014} and a 35k subset of validation images in COCO \textit{val2014}. Then, all ablation studies are conducted on the remaining 5k validation images (\texttt{minival}). To compare with the state-of-the-art models, we also submit our detection results to the COCO \texttt{test-dev} evaluation server. As the dataset uses average precisions (APs) at different IoUs and sizes as the main evaluation metrics, we report the standard COCO metrics including AP, AP$_{50}$, AP$_{75}$, and AP$_{S}$, AP$_{M}$, AP$_{L}$. 

\noindent\textbf{YOLOv3 with Residual Objectness.}
YOLOv3 already has an objectness module, but it suffers from an extreme foreground-background imbalance. To address the issue, we replace it by our residual objectness. In this case, while the step $T = 0$, the residual objectness is the same as the original objectness. We implement YOLOv3 with residual objectness (ReObj YOLOv3) by DarkNet~\cite{darknet}. In the original YOLOv3~\cite{yolov3}, the backbone network Darknet-53~\cite{yolov3} is pretrained on the ImageNet~\cite{imagenet} for detection network initialization. As the model is public available~\cite{darknet}, we simply initialize the network by it. 
All hyper-parameters are not changed~\footnote{\url{https://github.com/pjreddie/darknet/blob/master/cfg/yolov3.cfg}}. For example, we follow two original input scales (320$\times$320, 608$\times$608) in all experiments. Please refer to~\cite{yolov3,darknet} for more details.

\noindent\textbf{RetinaNet with Residual Objectness.}
RetinaNet~\cite{focal_loss} applies Focal Loss to down-weight the numerous easy negatives. 
To demonstrate that both residual objectness and Focal Loss can migrate the foreground-background/easy-hard imbalance, we abandon the usage of Focal Loss, but incorporate the residual objectness into RetinaNet (\textit{RetinaNet-ResObj}). 
To keep the consistency with~\cite{focal_loss}, we follow their hyper-parameter setting of ResNet-50-FPN~\footnote{\raggedright\url{https://github.com/facebookresearch/Detectron/blob/master/configs/12_2017_baselines/retinanet_R-50-FPN_1x.yaml}} and ResNet-101-FPN~\footnote{\raggedright\url{https://github.com/facebookresearch/Detectron/blob/master/configs/12_2017_baselines/retinanet_R-101-FPN_1x.yaml}} in principle, but use a 600 pixel train and test image scale in all ablation studies. Our implementation is based on \texttt{Detectron}~\cite{detectron}. 

\noindent\textbf{Faster R-CNN with Residual Objectness.}
In Faster R-CNN~\cite{faster_rcnn}, we apply our residual objectness for its region proposal network (RPN)~\cite{faster_rcnn}.
We perform our experiments on the \texttt{maskrcnn-benchmark}~\cite{maskrcnn_benchmark} with the hyper-parameters in public network configuration files for ResNet-50-FPN~\footnote{\raggedright\url{https://github.com/facebookresearch/maskrcnn-benchmark/blob/master/configs/e2e_faster_rcnn_R_50_FPN_1x.yaml}} and ResNet-101-FPN~\footnote{\raggedright\url{https://github.com/facebookresearch/maskrcnn-benchmark/blob/master/configs/e2e_faster_rcnn_R_101_FPN_1x.yaml}}. 

\subsection{Ablations}~\label{section:ablations}

\begin{table*}
    \centering
    \scriptsize
    \subtable[Gradient flow from residual subnet to/not to objectness subnet.]
    {
        \begin{tabular}[t]{c|ccc}
            \hline
            Choice & AP & AP$_{50}$ & AP$_{75}$ \\ 
            \hline
            \textit{To Objectness}  & 34.5 & 53.9 & 36.6 \\ 
            \textit{Not to Objectness} & \textbf{35.2} & \textbf{54.3} & \textbf{37.6} \\ 
            \hline
        \end{tabular}
    }\label{table:gradient}
    \subtable[Build residual head on the top of objectness/class head.]{
        \begin{tabular}[t]{c|ccc}
            \hline
            Top & AP & AP$_{50}$ & AP$_{75}$ \\ 
            \hline
            Class Head & 35.0 & 54.0 & 37.5 \\ 
            Objectness Head & \textbf{35.2} & \textbf{54.3} & \textbf{37.6} \\ 
            \hline
        \end{tabular}
    }\label{table:build}
    \subtable[Step $T$ with inference speed (on a NVIDIA Titan X GPU).]
    {
        \begin{tabular}[t]{c|c|ccc}
            \hline
            Step & FPS & AP & AP$_{50}$ & AP$_{75}$  \\
            \hline
            - & \textbf{10.2} & 34.2 & 53.1 & 36.5 \\
            \hline
            $T = 0$ & 10.2 & 34.1 & 52.9 & 36.6 \\
            $T = 1$ & 10.1 & 35.2 & 54.3 & 37.6  \\
            \textbf{$T = 2$} & 10.1 & \textbf{35.4} & \textbf{54.5} & \textbf{38.0}  \\
            $T = 3$ & 10.0 & \textbf{35.4} & \textbf{54.5} & \textbf{38.0} \\
            \hline
        \end{tabular}
    }\label{table:step}
    \caption{Ablations for \textit{RetinaNet-ResObj}. The “-” in the last table means vanilla RetinaNet.}\label{table2}
\end{table*}

\noindent\textbf{Gradient Flow.}
\textit{RetinaNet-ResObj} decouples the gradients of residual subnet and objectness subnet (see Figure~\ref{figure3}) during training, which means the residual subnet is supposed to independently refine the objectness estimation. 
In Table~\ref{table:gradient}, we compare this (\textit{Not to Objectness}) to recover the gradient flow (\textit{To Objectness}).  Nevertheless, this alternative results in a severe loss in detection accuracy (0.7 points). 
It may suggest that once we allow the gradient from residual subnet to objectness subnet, there will be a part of repeated loss computed by objectness subnet that not be conducive to residual learning.

\noindent\textbf{Build Residual Head.}
Table~\ref{table:build} shows the performance with the different location residual head built. While building it on the top of objectness head, it performs 0.2 AP higher than on class head.

\noindent\textbf{Steps.}
How many residual objectness heads should be added? As presented in Table~\ref{table:step}, while applying one residual head, the detector gains a large improvement (1.1 AP). But it is stable at $T \geq 2$, which achieves impressive 1.3 AP improvement with simple hyper-parameters tuning.
Nevertheless, building these layers slows down the detetcor, but this change is not obvious (0.2 FPS). 

\subsection{Results}\label{section:results}

\begin{table*}
\centering
\tiny
\begin{tabular}{l|c|c|c|cc|ccc}
\hline
Detectors & Size & Backbone & AP & AP$_{50}$ & AP$_{75}$ & AP$_{S}$ & AP$_{M}$ & AP$_{L}$ \\
\hline
\textit{RetinaNet-FocalLoss} & $1333\times800$ & ResNet-50-FPN & 35.7 & 55.0 & 38.5 & 18.9 & 38.9 & 46.3 \\
\textit{RetinaNet-ResObj} & $1333\times800$ & ResNet-50-FPN & \textbf{37.0} & \textbf{56.4} & \textbf{39.5} & \textbf{19.9} & \textbf{39.3} & \textbf{47.5} \\
\hline 
\textit{RetinaNet-FocalLoss}$^*$ & $1333\times800$ & ResNet-101-FPN & 39.1 & 59.1 & 42.3 & 21.8 & 42.7 & 50.2 \\
\textit{RetinaNet-GHM}$^*$ & $1333\times800$ & ResNet-101-FPN & 39.9 & \textbf{60.8} & 42.5 & 20.3 & \textbf{43.6} & \textbf{54.1} \\
\textit{RetinaNet-ResObj}$^*$ & $1333\times800$ & ResNet-101-FPN & \textbf{40.1} & 60.0 & \textbf{43.3} & \textbf{22.5} & 43.3 & 51.6 \\
\hline
\textit{YOLOv3-Obj} & $320\times320$ & DarkNet-53 & 28.2 & \textbf{51.2} & 28.6 & 9.0 & 29.7 & 44.5 \\
\textit{YOLOv3-ResObj} & $320\times320$ & DarkNet-53 & \textbf{29.3} & 50.6 & \textbf{30.7} & \textbf{9.5} & \textbf{31.1} & \textbf{45.0} \\
\hline
\textit{YOLOv3-Obj} & $608\times608$ & DarkNet-53 & 33.0 & \textbf{57.9} & 34.4 & 18.3 & 35.4 & 41.9 \\
\textit{YOLOv3-ResObj} & $608\times608$ & DarkNet-53 & \textbf{34.1} & 57.5 & \textbf{36.2} & \textbf{19.0} & \textbf{37.5} & \textbf{42.3} \\
\hline
\textit{FasterRCNN-Sampling} & $1333\times800$ & ResNet-50-FPN & 37.2 & \textbf{59.3} & 40.3 & 21.3 & 39.5 & 46.9 \\
\textit{FasterRCNN-ResObj} & $1333\times800$ & ResNet-50-FPN & \textbf{38.4} & \textbf{59.3} & \textbf{41.7} & \textbf{22.2} & \textbf{40.8} & \textbf{48.1} \\
\hline
\textit{FasterRCNN-Sampling} & $1333\times800$ & ResNet-101-FPN & 39.3 & \textbf{61.4} & 42.7 & 22.1 & 41.9 & 50.1 \\
\textit{FasterRCNN-ResObj} & $1333\times800$ & ResNet-101-FPN & \textbf{40.4} & 61.3 & \textbf{44.3} & \textbf{23.1} & \textbf{43.3} & \textbf{51.2} \\
\hline
\end{tabular}
(The item with * means the tricks followed ~\cite{focal_loss}, which is training detector with scale jitter and for 1.5 longer.)
\caption{Applying residual objectness to RetinaNet, YOLOv3 and FocalLoss, to replace their original solution for addressing the imbalance. 
On COCO \texttt{test-dev}, our residual objectness always achieves better detection accuracy. }\label{table:results} 
\end{table*}\label{table:results}

\noindent\textbf{COCO \texttt{test-dev}.}
In Table~\ref{table:results}, we compare residual objectness to other popular methods for addressing the imbalance on COCO \texttt{test-dev}. 
For RetinaNet, we adopt Focal Loss~\cite{focal_loss} and GHM~\cite{ghm} to conduct comparison. As the GHM did not report results with ResNet-50-FPN backbone, we only show the detection accuracy of \textit{RetinaNet-GHM} with ResNet-101-FPN backbone.   
While for Yolov3~\cite{yolov3} and Faster R-CNN~\cite{faster_rcnn}, we present the performance of their vanilla models which have objectness module and random sampling to alleviate the imbalance, respectively. 
By default, input images are resized such that their scale (shorter edge) is 800 pixels. However, the original YOLOv3 has not been trained in this scale. Hence, we use the scales of $320\times320$ and $608\times608$ that it adopted to train and inference.

$\bullet$ Residual Objectness vs. FocalLoss: With ResNet-50-FPN backbone, the \textit{RetinaNet-ResObj} is 1.3 AP (relative 3.6\%) higher than \textit{RetinaNet-FL}. We observed that all AP metrics are improved $\geq$ 1.0 AP with residual objectness. While for ResNet-101-FPN, the \textit{RetinaNet-ResObj} achieves a state-of-the-art 40.1 AP.  

$\bullet$ Residual Objectness vs. GHM: GHM harmonizes the distribution of gradient norm to alleviate the foreground-background imbalance, as well as down-weight the outliers. Thus, at the confusing metric AP$_{50}$, it surpasses Focal Loss and us. However, at the strict metric AP$_{75}$, our mechanism is obviously higher than GHM, which yields the better performance in the overall AP.

% need explain
$\bullet$ Residual Objectness vs. Objectness Module: YOLOv3 has already applied an objectness module to alleviate the imbalance, but it suffers from an extreme imbalance. 
As present in Table~\ref{table:results}, it has 1.1 AP (relative 3.9\%) lower than us at both $320\times320$ and $608\times608$ scales. 
Similar to GHM, our mechanism achieves lower performance at AP$_{50}$ metric but higher at AP$_{75}$. 
 
$\bullet$ Residual Objectness vs. Random Sampling: With replacing random sampling by residual objectness in RPN, the \textit{FasterRCNN-ResObj} achieves 1.2 AP (relative 3.2\%) and 1.1 AP higher than \textit{FasterRCNN-Sampling}, which indicates that our mechanism can improve the quality of the proposal boxes.  

With the impressive improvement residual objectness achieved, we still want to highlight its advantages. Unlike hand-crafted reweighting/sampling schemes, our residual mechanism is a fully learning-based algorithm that avoids laborious hyper-parameters tuning. 
Compared with current RPN-like/objectness modules, it progressively addresses the imbalance without any reweighting/sampling schemes. 
Finally, it is generalized for both one-stage and region-based detectors that has been presented in Table~\ref{table:results}.

\begin{figure}[h] 
    \centering
    \subfigure[\textit{First refinement}]{ 
    \includegraphics[width=0.47\linewidth]{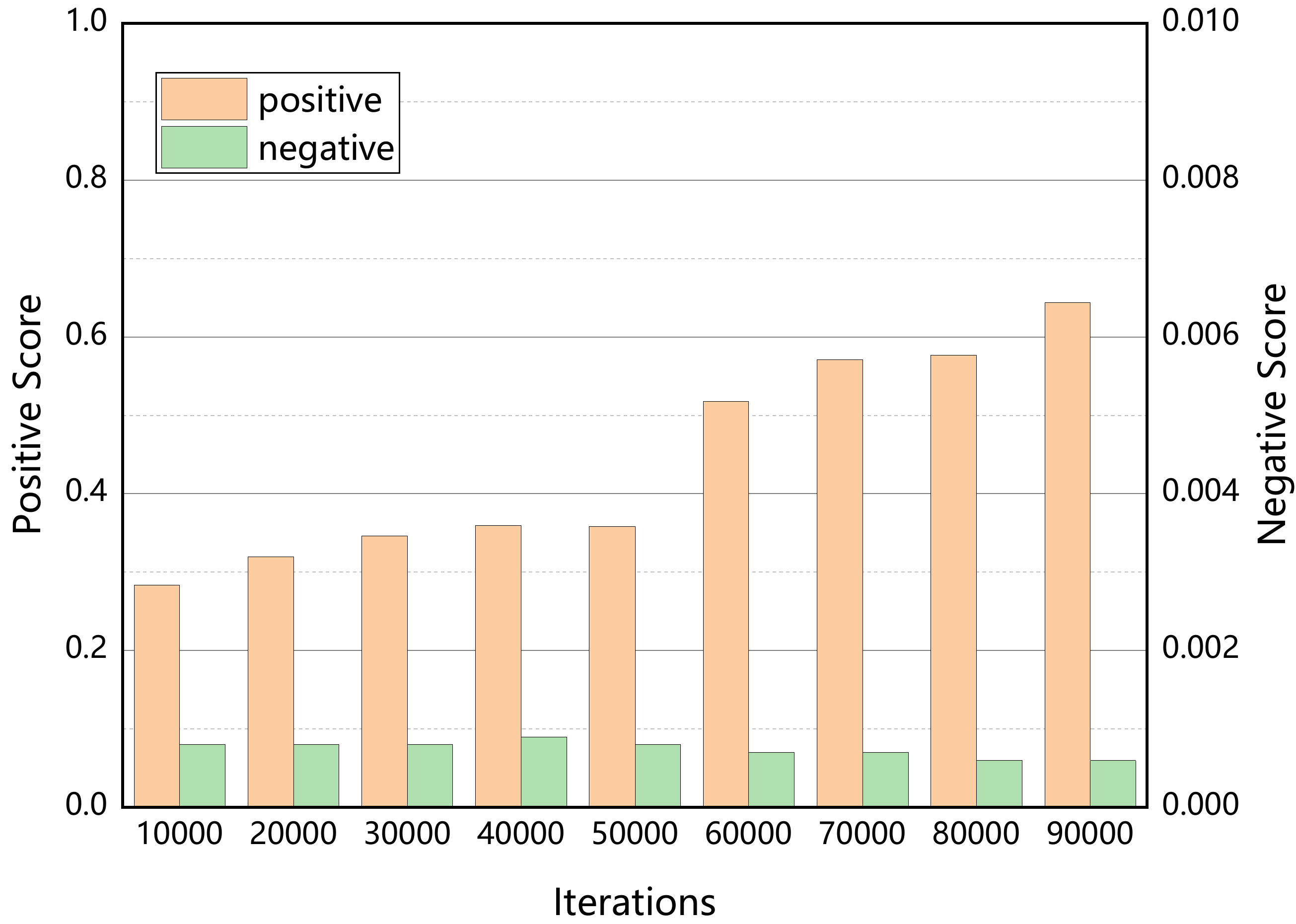}  
    \label{figure4a} 
    }
    \subfigure[\textit{Second refinement}]{
    \includegraphics[width=0.47\linewidth]{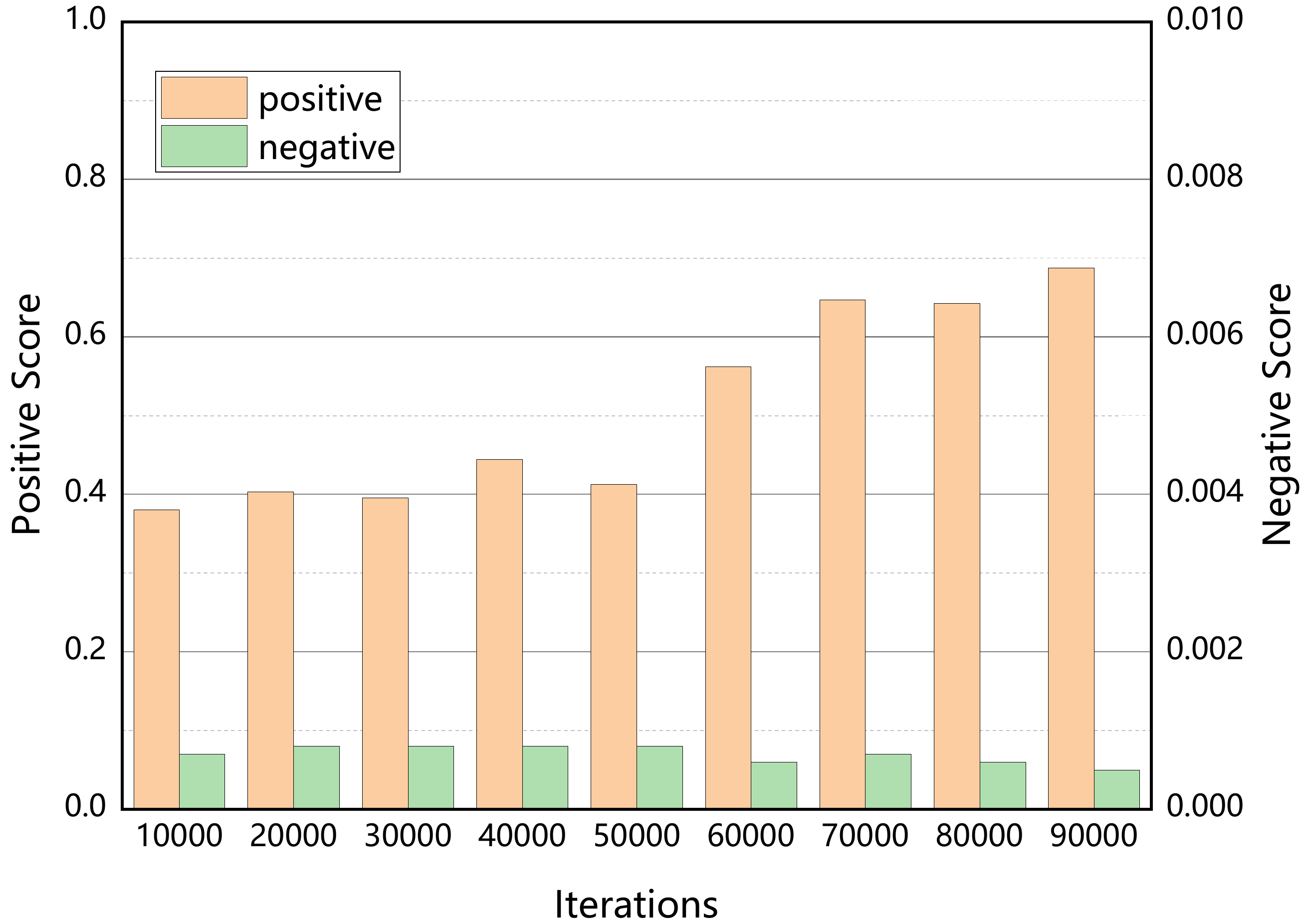}  
    \label{figure4b} 
    }
    \caption{Average objectness scores on COCO \texttt{minival} vary with training iterations. The results are achieved by \textit{RetinaNet-ResObj} of $T = 2$ model. }\label{figure4}
 \end{figure}

\noindent\textbf{Visualization.}
As shown in Figure~\ref{figure4}, we present the average objectness score of \textit{RetinaNet-ResObj} of $T = 2$ with ResNet-50-FPN~\cite{resnet,fpn} backbone,
which achieves the best detection accuracy in Section~\ref{section:ablations}.
Compared with Figure~\ref{figure2}, although the decrease in average negative objectness score is not clear due to its overwhelming number, the two refinements obviously improve the average positive objectness score, 
which helps better to better distinguish foregrounds from backgrounds. 

\begin{figure}[h] 
    \centering 
    \includegraphics[width=0.9\linewidth]{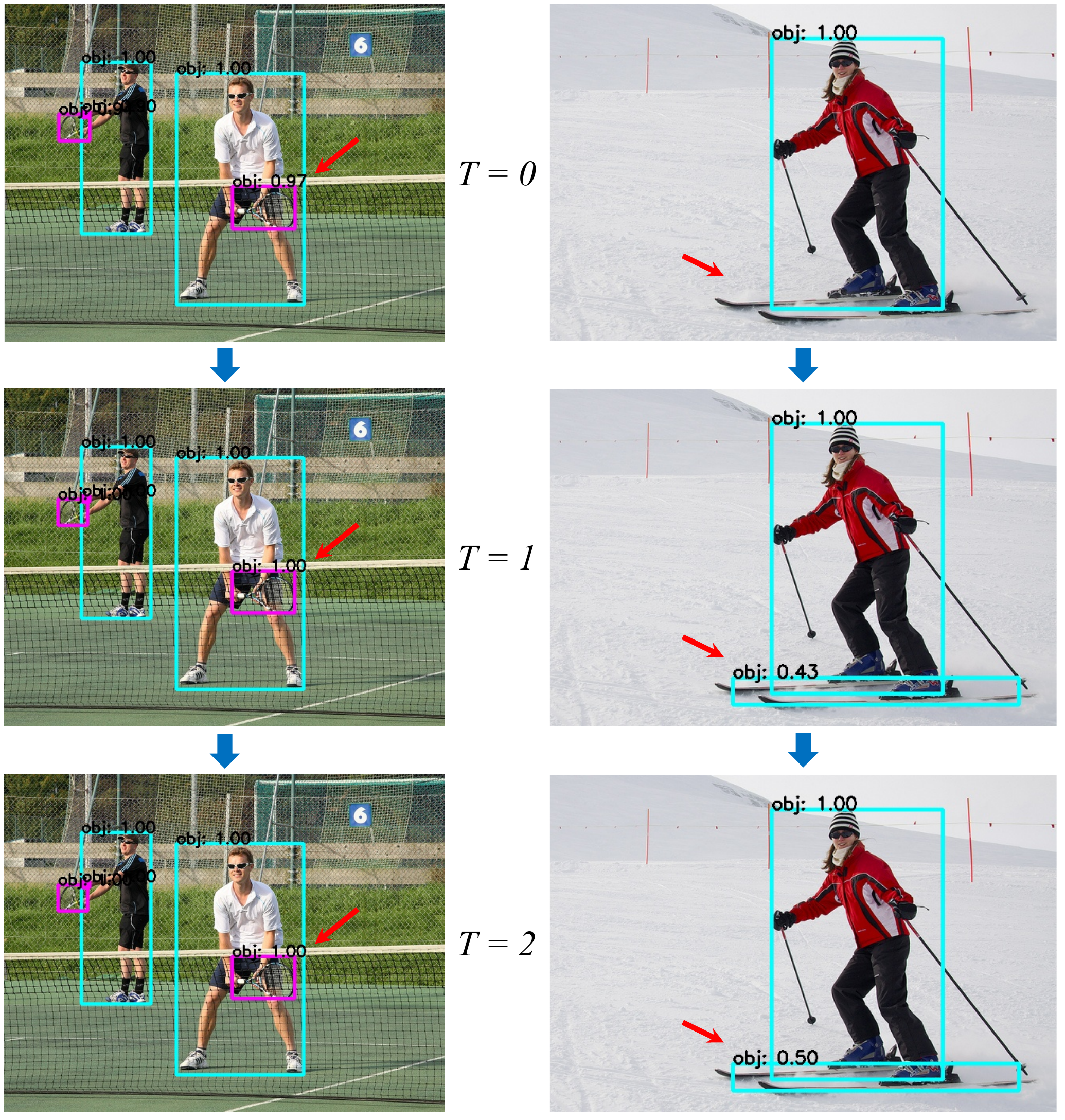} 
    \caption{Detection results (cases with score $> 0.4$ are shown) at different steps by \textit{RetinaNet-ResObj} with ResNet-50-FPN.
    It is explicitly shown (red array) that the object scores are gradually improved.} \label{figure5}
 \end{figure} 

To further illustrate our residual objectness mechanism, we visualize the detection bounding-boxes at $T=0$, $T=1$ and $T=2$ steps in inference. As shown in Figure~\ref{figure4}, the object scores can be gradually improved during the procedure, which helps to yield high-confidence predictions. 
In fact, our mechanism can also help to suppress the negative examples, but we do not visualize them here due to their tremendous quantity. 

\section{CONCLUSIONS}
In this paper, we investigated in object detectors, whether the heuristic sampling or reweighting schemes for addressing the imbalance are substitutable.
Our study surprisingly presented, with a simple learning-based objectness module rather than Focal Loss, the RetinaNet still achieved competitive detection accuracy.
Motivated by this, a fully learning-based mechanism termed \emph{Residual Objectness} was proposed, which helps detector to progressively address the imbalance.
Extensive experiments on COCO have demonstrated that for addressing the imbalance, the residual objectness is more effective than popular sampling/reweighting schemes.  
Given the state-of-the-art results on COCO, we expect the residual objectness would be adopted in various object detectors.

\bibliography{pr.bib}

\end{document}